\documentclass[final]{cvpr}

\usepackage{times}
\usepackage{epsfig}
\usepackage{graphicx}
\usepackage{amsmath}
\usepackage{amssymb}
\usepackage{multirow}
\usepackage{comment}

\usepackage[pagebackref=true,breaklinks=true,colorlinks,bookmarks=false]{hyperref}

\def\cvprPaperID{5421} 
\def\confYear{CVPR 2021}

\begin{document}

\title{Open-Vocabulary Object Detection Using Captions}

\author{
Alireza Zareian\textsuperscript{1,2},\; \;
Kevin Dela Rosa\textsuperscript{1},\; \;
Derek Hao Hu\textsuperscript{1},\; \;
Shih-Fu Chang\textsuperscript{2} \\
\textsuperscript{1} Snap Inc., Seattle, WA \;\;\;\;
\textsuperscript{2} Columbia University, New York, NY \\
{\tt\small \{azareian, kevin.delarosa, hao.hu\}@snap.com} \;\;\;\;
{\tt\small \{az2407, sc250\}@columbia.edu} \\
{\tt\small \href{https://github.com/alirezazareian/ovr-cnn}{github.com/alirezazareian/ovr-cnn}}
}

\maketitle

\begin{abstract}
    Despite the remarkable accuracy of deep neural networks in object detection, they are costly to train and scale due to supervision requirements. Particularly, learning more object categories typically requires proportionally more bounding box annotations. Weakly supervised and zero-shot learning techniques have been explored to scale object detectors to more categories with less supervision, but they have not been as successful and widely adopted as supervised models. In this paper, we put forth a novel formulation of the object detection problem, namely open-vocabulary object detection, which is more general, more practical, and more effective than weakly supervised and zero-shot approaches. We propose a new method to train object detectors using bounding box annotations for a limited set of object categories, as well as image-caption pairs that cover a larger variety of objects at a significantly lower cost. We show that the proposed method can detect and localize objects for which no bounding box annotation is provided during training, at a significantly higher accuracy than zero-shot approaches. Meanwhile, objects with bounding box annotation can be detected almost as accurately as supervised methods, which is significantly better than weakly supervised baselines. Accordingly, we establish a new state of the art for scalable object detection.
\end{abstract}
\section{Introduction}
\label{sec:intro}

\begin{figure}[t]
\begin{center}
 \includegraphics[width=1.00\linewidth]{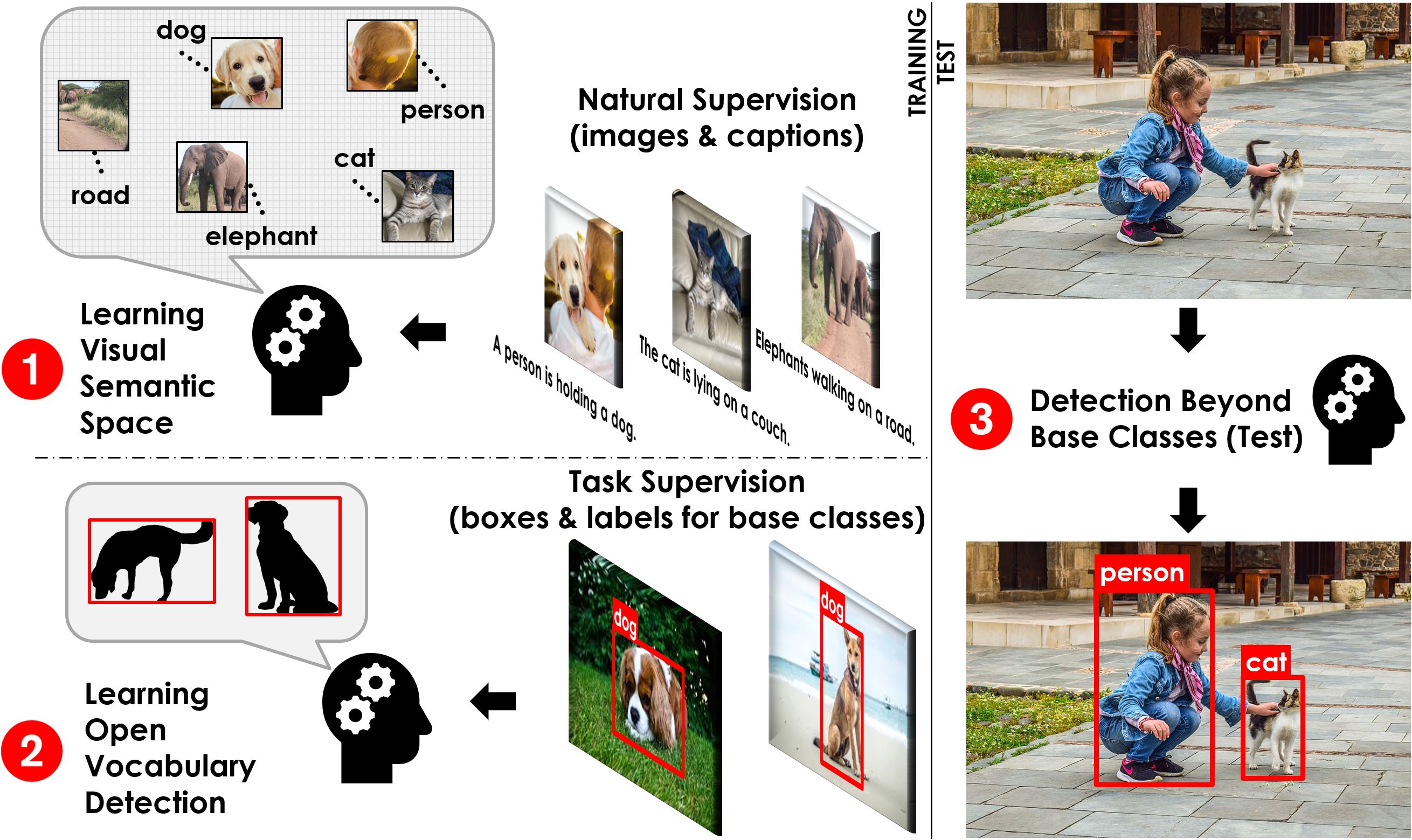}
\end{center}
   \caption{An overview of Open-Vocabulary Object Detection. We propose a two-stage training framework where we first (1) construct a visual-semantic space using low-cost image-caption pairs, and then (2) learn object detection using object annotations for a set of base classes. During test (3), the goal is to detect object categories beyond base classes, by exploiting the semantic space.}
\label{fig:ovd}
\end{figure}

Object detection is one of the most prominent applications of artificial intelligence, and one of the most successful tasks for deep neural networks. However, despite the tremendous progress in deep object detection, such as Faster R-CNN~\cite{ren2015faster} and its impressive accuracy, training such models requires expensive and time-consuming human supervision. Particularly, one needs to manually annotate at least thousands of bounding boxes for each object category of interest. Although such efforts have been already made and there are valuable datasets publicly available, such as Open Images~\cite{kuznetsova2020open} and MSCOCO~\cite{lin2014microsoft}, these datasets cover a limited set of object categories (\eg 600), despite requiring extensive resources. Extending object detection from 600 to 60,000 categories requires 100 times more resources, which makes versatile object detection out of reach.

Nevertheless, humans learn to recognize and localize objects effortlessly through natural supervision, \ie, exploring the visual world and listening to others describing situations. Their lifelong learning of visual patterns and associating them with spoken words results in a rich visual and semantic vocabulary that can be used not only for detecting objects, but for other tasks too, such as describing objects and reasoning about their attributes and affordances. Although drawing bounding boxes around objects is not a task that humans naturally learn, they can quickly learn it using few examples, and generalize it well to all types of objects, without needing examples for each object class.

In this paper, we imitate this human ability, by designing a two-stage framework named Open-Vocabulary object Detection (OVD). We propose to first use a corpus of image-caption pairs to acquire an unbounded vocabulary of concepts, simulating how humans learn by natural supervision, and then use that knowledge to learn object detection (or any other downstream task) using annotation for only some object categories. This way, costly annotation is only needed for some categories, and the rest can be learned using captions, which are much easier to collect, and in many cases freely available on the web~\cite{sharma2018conceptual}. Figure~\ref{fig:ovd} illustrates the proposed OVD framework, which is novel and efficient, enables versatile real-world applications, and can be generalized to other computer vision tasks.

\begin{figure}[t]
\begin{center}
 \includegraphics[width=1.00\linewidth]{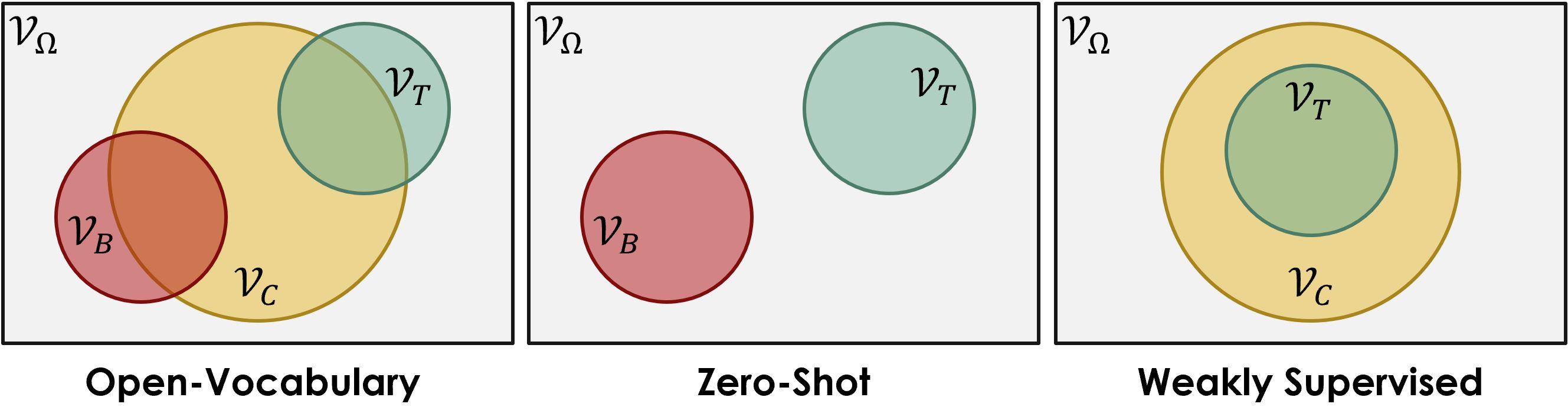}
\end{center}
   \caption{A comparison of our proposed OVD with existing ZSD and WSD paradigms. While zero-shot detection methods learn a limited set of base classes $\mathcal{V}_B$ and struggle to generalize to target classes $\mathcal{V}_T$, we acquire a much larger vocabulary $\mathcal{V}_C$ by learning from low-cost image-caption pairs. Although there are weakly supervised approaches that can learn from captions, they cannot use bounding box supervision from base classes, and need to know $\mathcal{V}_T$ before training. Hence, our OVD formulation is a generalization of ZSD and WSD, which can use both data sources to reach an outstanding performance on target classes not known in advance.}
\label{fig:venn}
\end{figure}

More specifically, we train a model that takes an image and detects any object within a given \textit{target} vocabulary $\mathcal{V}_T$. To train such a model, we use an image-caption dataset covering a large variety of words denoted as $\mathcal{V}_C$ as well as a much smaller dataset with localized object annotations from a set of \textit{base} classes $\mathcal{V}_B$. Note that in this task, target classes \underline{are not known} during training, and can be any subset of the entire language vocabulary $\mathcal{V}_\Omega$. This is in contrast with most existing object detection settings including weakly supervised transfer learning methods, where $\mathcal{V}_T$ should be known beforehand~\cite{uijlings2018revisiting}. The most similar task to OVD is zero-shot object detection, which also generalizes to any given target set, but cannot utilize captions. Figure~\ref{fig:venn} illustrates an intuitive abstraction of our proposed task compared to zero-shot and weakly supervised detection. Despite close connections to those well-known ideas, OVD is novel and uniquely positioned in the literature, as we elaborate in Section~\ref{sec:related}.

To address the task of OVD, we propose a novel method based on Faster R-CNN~\cite{ren2015faster}, which is first pretrained on an image-caption dataset, and then fine-tuned on a bounding box dataset, in a particular way that maintains the rich vocabulary learned during pretraining, enabling generalization to object categories without annotation. Through extensive experiments, we evaluate our method, Open Vocabulary R-CNN (OVR-CNN), and show that it achieves significantly higher performance than the state of the art in zero-shot learning (27\% mAP compared to 10\%). We also show that it outperforms weakly supervised object detectors by a significant margin in generalized zero-shot settings (40\% mAP compared to 26\%). We supplement the paper with comprehensive open-source code to reproduce results.\footnote{\url{https://github.com/alirezazareian/ovr-cnn}}

\section{Related work}
\label{sec:related}

\paragraph{Zero-shot object detection (ZSD)}
aims to generalize from annotated (seen) object classes to other (unseen) categories. 
The key idea is to use zero-shot learning techniques (\eg word embedding projection~\cite{frome2013devise}) to learn object proposal classification.
Bansal \etal~\cite{bansal2018zero} argued the main challenge in ZSD is modeling the background class, which is hard to separate from unseen classes.
They defined background as a mixture model, which was later improved by the introduction of polarity loss~\cite{rahman2020improved}.
On the other hand, Zhu \etal~\cite{zhu2019zero,zhu2020don} argued the key to ZSD is to improve the generalization ability of object proposal models. They employed a generative model to hallucinate unseen classes and augment seen examples when training the proposal model.
Nevertheless, ZSD methods are still far from practical performance, due to their unnecessarily harsh constraint, \ie, not having any example of unseen objects, and having to guess how they look like solely based on their word embeddings~\cite{bansal2018zero,rahman2020improved,zhu2020don} or textual descriptions~\cite{li2019zero}. This has motivated recent papers to simplify the task by making unrealistic assumptions, such as the availability of test data during training~\cite{rahman2019transductive}, or the availability of unseen class annotations to filter images with unseen object instances~\cite{gupta2020multi}. Considering datasets with natural, weak supervision are abundant and cheap, we propose an alternative, more realistic problem: Besides annotated data for ``seen'' classes, we assume an image-caption dataset is available that covers a larger variety of objects with an open vocabulary. 
This allows us to achieve 27\% mAP on unseen classes, compared to the 10\% state of the art, without much extra annotation effort. To this end, we address the open problem of knowledge transfer from image-caption pretraining to object detection.

\vspace{-0.4cm}
\paragraph{Weakly supervised object detection (WSD)} is the most widely used approach to train object detectors without bounding box annotations, by using image-level labels instead. The main challenge of WSD is localization, as each label may refer to any object in the image. This problem is typically addressed using multiple instance learning, which is a well-studied topic~\cite{bilen2016weakly,wan2019c,cinbis2016weakly}. Although image-level labels are easier to collect than bounding boxes, they still require manual effort, and they are typically limited to a predefined taxonomy. In contrast, we use captions, which are more natural to annotate and often freely available on the web, while also featuring a rich, open vocabulary of concepts. Learning object detection from captions has been studied at a limited scale. Cap2Det~\cite{ye2019cap2det} parses captions into multi-label classification targets, which can be used to train a WSD model. However, that requires image-level labels to train the caption parser, and is limited to a closed vocabulary. Amrani \etal~\cite{amrani2019learning} train a WSD model based on the presence of a predefined set of words in captions, which is similarly closed-vocabulary, and discards the rich semantic content of captions, which we exploit through transformers. In contrast, Sun \etal~\cite{sun2015automatic} and Ye \etal~\cite{ye2018learning} aim to discover an open set of object classes from image-caption corpora, and learn detectors for each discovered class. A key limitation of all such WSD methods is their inferior object localization accuracy. In contrast, we disentangle object recognition and localization into two independent problems. We learn recognition using open-vocabulary captions, while learning localization using a fully annotated dataset from a small subset of classes. 

\vspace{-0.3cm}
\paragraph{Object detection using mixed supervision} has been studied in order to exploit both weak and full supervision. However, most existing methods need bounding box annotations for all classes, and use weak supervision only as auxiliary data~\cite{gao2019note,ramanathan2020dlwl,wang2015model}. More similar to our work are those which transfer a detector trained on supervised base classes to weakly supervised target classes~\cite{hoffman2014lsda,tang2016large,uijlings2018revisiting}. These methods usually lose performance on base classes as we show in Section~\ref{sec:exp}. In contrast, we treat this problem as an opposite knowledge transfer process: Instead of training on base classes first, and transferring to target classes using weakly supervised learning, we first use captions to learn an open-vocabulary semantic space that includes target classes, and transfer that to the task of object detection via supervised learning. Another limitation of all weakly supervised and mixed-supervision methods is that they require image-level annotations within a predefined taxonomy, and they only learn those predefined classes. In contrast, we use captions which are open-vocabulary and also more prevalent on the web, and we learn to generalize to any set of target classes on demand, without having to know them beforehand. VirTex~\cite{desai2020virtex} is the only method that uses captions as well as object annotations to train a detector, but it needs annotation for all object classes while we can generalize from a subset of annotated categories.

\vspace{-0.3cm}
\paragraph{Visual grounding of referring expressions}
can be seen as an open-vocabulary object localization problem: Given an image and a noun phrase that refers to an object, usually within the context of a full caption, the goal is to localize the referred object in the image using a bounding box. We are inspired by the rich literature of weakly supervised visual grounding methods~\cite{xiao2017weakly,datta2019align2ground,chen2018knowledge,akbari2019multi} to design our image-caption pretraining technique. More specifically, we learn to map caption words to image regions, by learning a visual-semantic common space. However, such a mapping alone cannot be used to detect objects during inference when no caption is provided. Therefore, we propose to transfer visual grounding knowledge to the task of object detection through another phase of training.

\vspace{-0.2cm}
\paragraph{Vision-language transformers}
Our framework of pretraining with image-captions and transferring the learned knowledge to the downstream task is inspired by the recent success of multimodal transformers~\cite{lu2019vilbert,li2019visualbert,li2020weakly,chen2020uniter,su2020vlbert}. These methods train transformers in a self-supervised manner to take image-caption pairs as input and extract versatile features that can be fine-tuned on various downstream vision-language tasks. However, they have not been applied to object detection yet, since they need both image and caption as input, and also because they rely on a pretrained object detector to articulate the image before feeding into transformers. Recently, PixelBERT~\cite{huang2020pixel} removed the latter requirement by applying transformers directly on the feature map. We utilize and extend PixelBERT by devising a vision-to-language projection module before feeding visual features into the transformer, and by incorporating a visual grounding cost function into the pretraining process. Moreover, we propose to transfer the weights pretrained via multi-modality transformers to the single-modality downstream task of object detection.
\section{Method}
\label{sec:method}

\begin{figure*}[th]
\begin{center}
\includegraphics[width=0.8\linewidth]{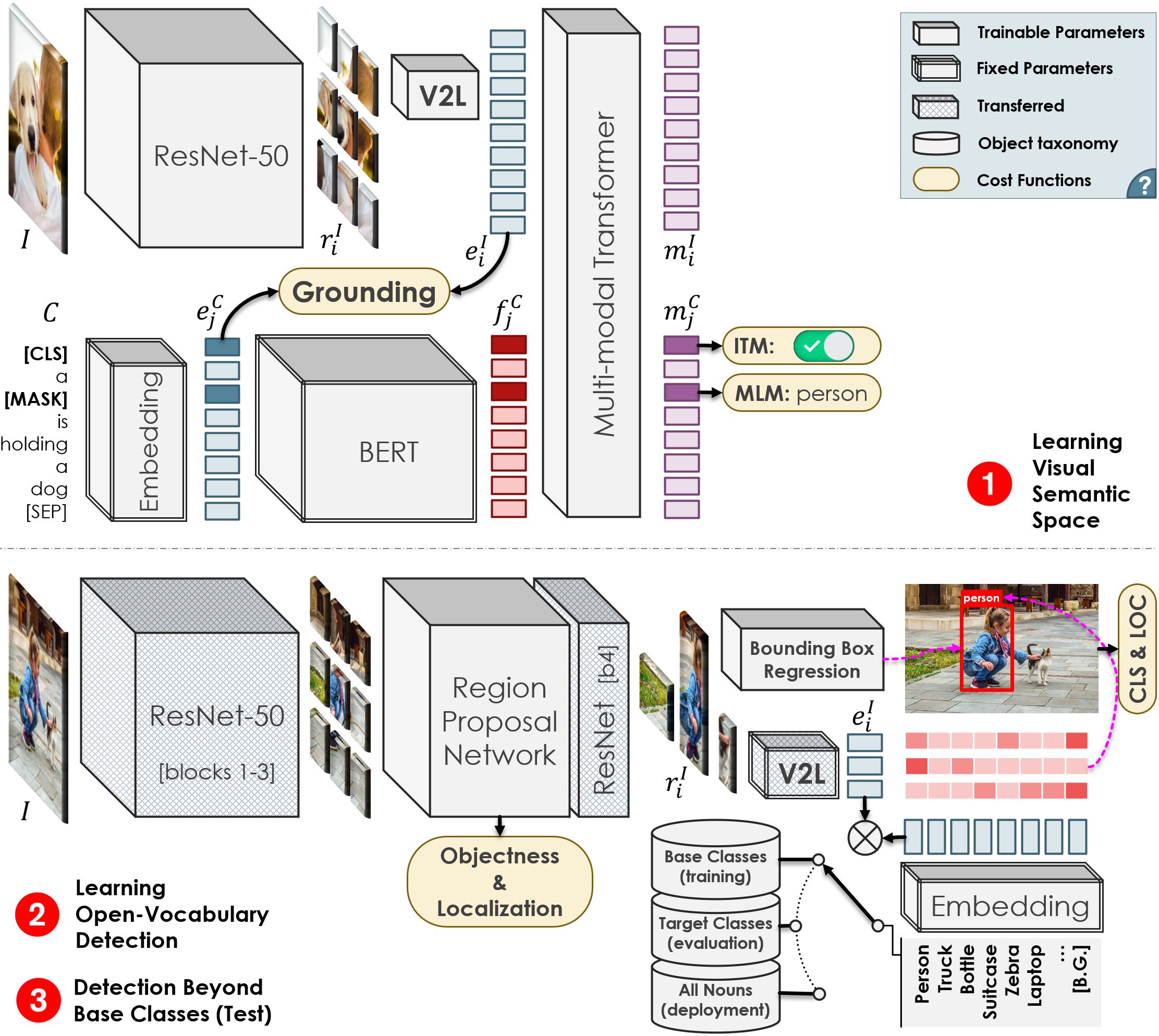}
\end{center}
\caption{The architecture of our OVR-CNN during pretraining (top) and downstream training (bottom). We first train the ResNet and the V2L layer on image-caption pairs via grounding, masked language modeling (MLM) and image-text matching (ITM). Then we use the trained ResNet and V2L to initialize a Faster R-CNN in order to learn open-vocabulary object detection.}
\label{fig:ovrcnn}
\end{figure*}

Figure~\ref{fig:ovrcnn} illustrates the architecture of our proposed method, which is based on a Faster R-CNN~\cite{ren2015faster} trained in a zero-shot manner. More specifically, it is trained on a set of \textit{base} classes $\mathcal{V}_B$, and tested on another set of \textit{target} classes $\mathcal{V}_T$. 
To this end, pretrained word embeddings (\eg GloVE~\cite{pennington2014glove}) are often used instead of conventional, trainable classifiers, so that target class embeddings can replace base class embeddings during testing, without changing the model's output semantic space. Nevertheless, this practice often leads to severe overfitting due to the small sample of base classes, to the point where the state-of-the-art mAP on target classes is 9 times lower than base classes~\cite{rahman2020improved}. 

To alleviate this problem, our key idea is to pretrain the visual backbone on a larger vocabulary $\mathcal{V}_C$ to learn a more complete semantic space rather than a small number of base classes. Since captions are naturally written without much constraint on the vocabulary, they are a perfect source for learning a rich and complete visual-semantic space. We name this framework Open Vocabulary Object Detection (OVD), as there are no explicit limits on the vocabulary of objects that can be learned through captions. In practice, our vocabulary is not literally ``open'', as it is limited to pretrained word embeddings. However, word embeddings are typically trained on very large text corpora such as Wikipedia that cover nearly every word~\cite{pennington2014glove,devlin2018bert}. 

In the rest of this section, we elaborate how we pretrain our Open Vocabulary faster R-CNN (OVR-CNN) on image-caption pairs, and how we transfer the pretraining knowledge to the downstream task. In Section~\ref{sec:exp}, we demonstrate that our method closes the base-target performance gap from a ratio of 9 to 2.

\subsection{Learning a visual-semantic space}

Object detectors typically use a CNN backbone that is often pretrained for ImageNet classification~\cite{deng2009imagenet,ren2015faster}. Pretraining results in a backbone that can extract features optimized for object recognition, which is then used to train a new classification head for a fixed set of annotated classes. This is problematic in zero-shot settings, as a classifier trained on base classes cannot recognize target classes. Therefore, zero-shot methods learn a linear projection from visual features to pretrained base class embeddings by replacing classifier weights with a fixed embeddings matrix~\cite{frome2013devise}. This way, the network is expected to generalize to target classes by assuming the continuity of the embedding space. Nevertheless, this approach is prone to overfitting, as projecting to a small number of the embedding space (base class embeddings) is an under-determined problem~\cite{bansal2018zero}. 

To prevent overfitting, we propose to learn the aforementioned Vision to Language (V2L) projection layer along with the CNN backbone during pretraining, where the data is not limited to a small set of base classes. 
To this end, we use an image-caption dataset, since captions contain a rich vocabulary and semantic structure that can be used to learn the meaning of words, including object names. To effectively learn from the rich supervision that captions provide, we exploit recent advances in visual grounding and vision-language transformers. We use a main (grounding) task as well as a set of auxiliary self-supervision tasks to learn a robust CNN backbone and V2L layer. In the next subsection, we elaborate how we transfer the pretrained modules to learn open-vocabulary object detection.

Our pretraining architecture resembles PixelBERT~\cite{huang2020pixel}: it takes image-caption pairs as input, feeds the image into a visual backbone and the caption into a language backbone, which results in a set of token embeddings for the image and caption, and then feeds those token embeddings into a multimodal transformer to extract multimodal embeddings. Our visual backbone is a ResNet-50~\cite{he2016deep}, which takes a $w \times h$ image $I$ as input and extracts a grid of $w/32 \times h/32$ regions, where each region $i$ is represented by a $d_v$-dimensional feature vector, $r_i^I$. For the language backbone, we use a pretrained BERT~\cite{devlin2018bert}, which takes a tokenized caption $C$ as input, extracts a $d_l$-dimensional word embedding $e_j^C$ for each token $j$, augments that with position embeddings, and applies several layers of multi-head self-attention to extract $d_l$-dimensional contextualized token embeddings $f_j^C$. 

Furthermore, we devise a linear V2L layer that maps each visual region representation $r_i^I$ into the language embedding space $e_i^I$.
The final embeddings of image regions $\{e_i^I\}$ and caption tokens $\{f_j^C\}$ are then aggregated and fed into a multimodal transformer, which is similar to BERT in architecture, but performs attention not only within each modality but also across the two modelities. The output of the multimodal transformer is $\{m_i^I\}$ and $\{m_j^C\}$ for the regions and words respectively, which can be used for various pretraining tasks, as we discuss later in this section.


Once we extract the aforementioned stages of unimodal and multimodal embeddings from a batch of image-caption pairs, we define a main objective function as well as various auxiliary objectives to ensure an effective training for the ResNet parameters, as well as the V2L layer. Our main objective is visual grounding, \ie, word embeddings from each caption $e_j^C$ should be close to their corresponding image regions $e_i^I$. Since the correspondence of words and regions is not given, we employ a weakly supervised grounding technique to learn it.
Specifically, we define a global grounding score for each image-caption pair, that is a weighted average of local grounding scores for word-region pairs:
\begin{equation}
\begin{aligned}
\langle I,C \rangle_G = \frac{1}{n_C} \sum_{j=1}^{n_C} \sum_{i=1}^{n_I} a_{i,j} \langle e_i^I, e_j^C \rangle_L
,\end{aligned}
\end{equation}
where $\langle ., . \rangle_L$ is the dot product of two vectors, $n_I$ and $n_C$ are the number of image and caption tokens, and
\begin{equation}
\begin{aligned}
a_{i,j} = \frac{\exp \langle e_i^I, e_j^C \rangle_L}{\sum_{i'=1}^{n_I} \exp \langle e_{i'}^I, e_j^C \rangle_L} 
.\end{aligned}
\end{equation}
The global grounding score for a matching image-caption pair should be maximized, while it should be minimized for a non-matching pair. Hence, we use other images in the batch as negative examples for each caption, and use other captions in the batch as negative examples for each image. Accordingly, we define two grounding objective functions:
\begin{equation}
\begin{aligned}
\mathcal{L}_G(I) = - \log \frac{\exp \langle I, C \rangle_G}{\sum_{C' \in \mathcal{B}_C} \exp \langle I, C' \rangle_G}
,\end{aligned}
\end{equation}
and
\begin{equation}
\begin{aligned}
\mathcal{L}_G(C) = - \log \frac{\exp \langle I, C \rangle_G}{\sum_{I' \in \mathcal{B}_I} \exp \langle I', C \rangle_G}
,\end{aligned}
\end{equation}
where $\mathcal{B}_I$ and $\mathcal{B}_C$ are the image and caption batch.
We validated the described formulation by completing extensive experimentation with various other alternatives, such as other similarity metrics (\eg cosine instead of dot product), other loss functions (\eg triplet loss instead of negative log likelihood) and other word-to-region alignment mechanisms (\eg hard alignment instead of softmax). 

Optimizing the grounding objectives results in a learned visual backbone and V2L layer that can map regions in the image into words that best describe them, without limiting to a closed vocabulary. However, since we induce a weak, indirect supervision, a local optima might be achieved where the model only learns the minimum concepts necessary to choose the right image/caption. To more directly learn each word, we employ masked language modeling following PixelBERT~\cite{huang2020pixel}. Specifically, we randomly replace some words $j$ in each caption $C$ with a [MASK] token, and try to use the multimodal embedding of the masked token $m_j^C$ to guess the word that was masked. To this end, the visual backbone and the V2L layer should learn to extract all objects that might be described in captions, and the multimodal transformer should learn to use those along with the language understanding ability of BERT to determine what word completes the caption best. 

Accordingly, we apply a fully connected layer on $m_j^C$, compare its output to all word embeddings using dot product, and apply softmax to compute a probability score for each word. We define masked language modeling $\mathcal{L}_{MLM}$ as a cross-entropy loss comparing the predicted distribution with the actual word that was masked. PixelBERT also employs an image-text matching loss $\mathcal{L}_{ITM}$, but does not use masked visual modeling that is common in vision-language transformers~\cite{lu2019vilbert}. We follow their choices for our auxiliary objectives, although other combinations are possible. We train the visual backbone, V2L layer, and the multimedia transformer jointly by minimizing the total loss for each image-caption pair:
\begin{equation}
\begin{aligned}
\mathcal{L}(I,C) = \mathcal{L}_G(I) + \mathcal{L}_G(C) + \mathcal{L}_{MLM} + \mathcal{L}_{ITM}
.\end{aligned}
\end{equation}
Note that our language backbone (BERT) and its word embeddings are fixed in our experiments.

\subsection{Learning open-vocabulary detection}

Once the ResNet visual backbone and V2L layer are trained, we transfer them to the task of object detection, by initializing a Faster R-CNN. Following~\cite{ren2015faster}, we use the stem and the first 3 blocks of our pretrained ResNet to extract a feature map from a given image. Next, a region proposal network slides anchor boxes on the feature map to predict objectness scores and bounding box coordinates, followed by non-max suppression and region-of-interest pooling to get a feature map for each potential object. Finally, following~\cite{ren2015faster}, the 4th block of our pretrained ResNet is applied on each proposal followed by pooling to get a final feature vector $r^I_i$ for each proposal box, which is typically fed into a classifier in supervised settings. 

Nevertheless, in our zero-shot setting, a linear layer is applied on the visual features $r^I_i$ to map each proposal onto a word embedding space $e^I_i$, so they can be compared to base or target class embeddings in the training or testing phase respectively. In all ZSD methods, the aforementioned linear layer is trained from scratch on base classes, which struggles to generalize. In contrast, we have already trained the V2L layer in the pretraining phase, on a much broader semantic space. The main difference of this phase with pretraining is that instead of the grid-structured feature map, $r^I_i$ represents a bounding box of arbitrary shape. However, due to the linear characteristics of RoI-Align~\cite{he2017mask}, $r^I_i$ is on the same space as in pretraining, with minimal domain shift that can be eliminated by fine-tuning the ResNet backbone.

During training, we compare $e^I_i$ to each base class $k$ to compute classification scores:
\begin{equation}
\begin{aligned}
p(i\text{ classified as }k) = \frac{\exp \langle e^I_i, e^\mathcal{V}_k \rangle}{1 + \sum_{k'\in\mathcal{V}_B} \exp \langle e^I_i, e^\mathcal{V}_{k'} \rangle}
,\end{aligned}
\label{eq:classification}
\end{equation}
where $e^\mathcal{V}_k$ is the pretrained embedding of word $k$, $\mathcal{V}_B$ is the set of base classes, and $\langle .,. \rangle$ denotes dot product. The addition of 1 in the denominator is because we set the background class to have a fixed, all-zero embedding, which makes any dot product zero, and is exponentiated to 1. We found that a fixed all-zero background embedding performs better than a trainable one as it does not push non-foreground bounding boxes, which may contain target classes, to an arbitrary region of the embedding space.

Except for the aforementioned changes in the classification head, the rest of our network exactly follows Faster R-CNN, and is trained in the exact same way with the same objective functions. Empirically, we found that multiplying a ratio $\alpha$ to the classification loss of background proposals (\ie, proposal boxes that are not matched with any ground truth bounding box) can improve the performance on target classes significantly, while slightly lowering base class performance. Hence, we use cross-validation to find the best $\alpha$ for each model. The ResNet parameters are finetuned, while the region proposal network and the regression head are trained from scratch. The classifier head is fully fixed, as it consists of a pretrained V2L layer and word embeddings, which are especially prone to overfitting. During testing, we use the model just like a Faster R-CNN, except we can replace word embeddings in Eq.~(\ref{eq:classification}) with any set of target classes $\mathcal{V}_T$. While we evaluate on a fixed, annotated target set, the model is not particularly tuned for those classes, and hence can be deployed on the entire vocabulary $\mathcal{V}_\Omega$.

\section{Experiments}
\label{sec:exp}

In this section, we demonstrate our method's ability to detect objects of the target classes accurately, while not losing its accuracy on the base classes compared to supervised approaches. Particularly, we show significant quantitative improvements compared to zero-shot and weakly supervised object detection methods, followed by a comprehensive analysis including ablation and visualization. 

\subsection{Data and metrics}
We base our experiments on the challenging and widely used COCO Objects dataset~\cite{lin2014microsoft}. We use their 2017 training and validation split for training and evaluation respectively. To select base and target classes, we adopt the split proposed by~\cite{bansal2018zero} and used by all other ZSD methods. Their splits includes 48 base classes and 17 target classes, which are both subsets of COCO object classes. We remove any bounding box that is not labeled with a base class from training data, and remove images that are left with no bounding boxes. This leaves us with 107,761 training images that contain 665,387 instances of base classes, and 4,836 test images that contain 28,538 instances of base classes and 4,614 instances of target classes.

Unless otherwise mentioned, for pretraining we use COCO Captions~\cite{chen2015microsoft}, which is based on the same images and same train/test split as COCO Objects. This dataset is preferred due to the matching domain with the downstream task. However, to study more realistic settings, we also report results by pretraining on Conceptual Captions (CC)~\cite{sharma2018conceptual}, which was automatically collected from the web.
CC is larger with 2,749,293 training image-caption pairs, compared to COCO with 118,287 images and 5x captions. Both COCO and CC cover very broad vocabularies that include all base and target classes in our experiments. Although there is no theoretical limit for our model to predict words outside the caption vocabulary $\mathcal{V}_C$ (Figure~\ref{fig:venn}), we do not study those unusual cases in this paper.

Following most ZSD and WSD methods, we evaluate using mean Average Precision (mAP) at an IoU of 0.5. We compute mAP on base classes by directly applying the model on COCO validation images and using base class annotations to evaluate. Then we replace the classifier head with target class embeddings and apply on the same dataset, but this time compare with target class annotations. These result in base and target mAP, which resemble supervised and zero-shot settings respectively. We also replace the classifier head with the union of base and target class embeddings, to mimic generalized zero-shot settings~\cite{rahman2020improved}. In that case, we report total mAP, as well as separately computing the mean of AP over base and target classes. 

\subsection{Implementation details}

We used the \texttt{maskrcnn-benchmark} code base~\cite{massa2018mrcnn}, and particularly the \texttt{R\_50\_C4} configuration to implement our system. We also used a pretrained and frozen \texttt{BERT-Base}~\cite{Wolf2019HuggingFacesTS} as our language backbone. For the multimodal transformer, we use the same architecture as \texttt{BERT-Base}, except we use only 6 layers and 8 attention heads at each layer, and we train it from scratch. Our base learning rate for pretraining is 0.01 which drops to 0.001 and 0.0001 after sufficient training. We use a batch size of 64 and train on 8 V-100 GPUs which takes about 10 hours. We use spatial dropout following~\cite{huang2020pixel} to subsample visual regions during pretraining. For masked language modeling, we mask each word with the likelihood of 0.135. We use gradient clipping at 5.0 for pretraining.

During downstream training, we use the BERT embeddings (\ie, pretrained input embeddings, not the output of BERT transformers) of the base classes to initialize and fix the classifier weights. We found the best background weight is $\alpha=0.2$ for most experiments, except the ablations without a fixed, pretrained V2L layer, where $\alpha=0.0$ works best. We only fine-tune the third and forth block of ResNet, and keep the stem and first two blocks fixed. We train using a learning rate of 0.005 and drop to 0.0005 and 0.00005 when appropriate. We train with a batch size of 8 on 8 V-100 GPUs which takes about 18 hours to converge.

\subsection{Baselines}

Because our proposed OVD task utilizes a unique combination of training data and supervision that has not been studied before, there are no baselines with identical training settings for an entirely fair comparison. Therefore, we use baselines from a variety of similar but not identical tasks.
Firstly, we compare to zero-shot detection methods, as ZSD is the closest area to our work. Particularly, we compare to SB~\cite{bansal2018zero}, which is the first and simplest ZSD method, projecting CNN features of EdgeBox proposals~\cite{zitnick2014edge} to word embeddings. Then we compare to LAB~\cite{bansal2018zero}, which attempts to better model the background class using a mixture model. We also compare to DSES~\cite{bansal2018zero}, which uses additional classes from Visual Genome~\cite{krishna2017visual} to augment base classes. Then we compare to PL~\cite{rahman2020improved}, which proposes polarity loss to address the object-background imbalance, and to DELO~\cite{zhu2020don}, which employs a generative approach to prepare the model for certain target classes through feature hallucination. Note that DELO needs to know target classes beforehand, which makes it not truly open-vocabulary.

It is important to note that our approach utilizes extra data (COCO Captions or Conceptual Captions) that is not available to ZSD baselines, and may include examples of target categories. Therefore, we also compare to weakly supervised detection (WSD) methods, by converting captions into image-level labels using exact matching or a classifier~\cite{ye2019cap2det}. We compare to WSDDN~\cite{bilen2016weakly}, as well as Cap2Det~\cite{ye2019cap2det} which better utilizes captions. WSD methods do not utilize bounding boxes for base classes, which can be an advantage in situations where no such annotation is available, but it results in poor localization performance compared to our method, which is able to utilize bounding boxes. Hence, we also compare to transfer learning methods that utilize a mixture of weak and full supervision (denoted as MSD). Particularly, we compare to LSDA~\cite{hoffman2014lsda}, which learns a transformation from classifier weights into detector weights, its extension~\cite{tang2016large} to utilize semantic class relationships (LSDA+), and a more recent work~\cite{uijlings2018revisiting} which uses multiple instance learning along with a region proposal network trained on base classes (MIL+RPN).

Note that since WSD and MSD methods require image-level labels, target classes should be known in advance during pretraining, and the models are particularly adapted to those classes. In contrast, our method and most ZSD methods have no access to such information, and can be applied to any novel class without retraining.

\subsection{Results}

\begin{table}[tb]
\begin{center}
\footnotesize
\begin{tabular}{l|c|c|c|c c c}
\hline
\multirow{2}{*}{Method} & \multirow{2}{*}{Task} & Base & Target & \multicolumn{3}{|c}{Generalized (48+17)} \\
&& (48) & (17) & Base & Target & All \\ 
\hline
\hline
FR-CNN \cite{ren2015faster} & FSD & 54.5 & - & - & - & -   \\

\hline
WSDDN \cite{bilen2016weakly}* & \multirow{2}{*}{WSD} & - & - & 19.6 & 19.7 & 19.6 \\
Cap2Det \cite{ye2019cap2det}* && - & - & 20.1 & 20.3 & 20.1 \\

\hline
LSDA \cite{hoffman2014lsda}* & \multirow{3}{*}{MSD} & - & - & 29.3 & 17.7 & 27.2 \\
LSDA+\cite{tang2016large}* && - & - & 28.5 & 21.9 & 26.7 \\
MIL+RPN\cite{uijlings2018revisiting}* && - & - & 27.8 & 22.6 & 26.4 \\

\hline

SB \cite{bansal2018zero} & \multirow{5}{*}{ZSD} & 29.7 & 0.70 & 29.2 & 0.31 & 24.9  \\
LAB \cite{bansal2018zero} && 21.1 & 0.27 & 20.8 & 0.22 & 18.0  \\
DSES \cite{bansal2018zero} && 27.2 & 0.54 & 26.7 & 0.27 & 22.1  \\
DELO \cite{zhu2020don}* && 14.0 & 7.60 & 13.8 & 3.41 & 13.0 \\
PL \cite{rahman2020improved} && 36.8 & 10.0 & 35.9 & 4.12 & 27.9  \\

\hline
\textbf{OVR-CNN} & \textbf{OVD} & \textbf{46.8} & \textbf{27.5} & \textbf{46.0} & \textbf{22.8} & \textbf{39.9} \\
\hline
\end{tabular}
\end{center}
\caption{Results on the MSCOCO dataset. Numbers are mAP (\%). *For some baselines, target classes are known during training.}
\label{table:main_results}
\vspace{-0.1cm}
\end{table}

Table~\ref{table:main_results} demonstrates our main results compared to the baselines. Particularly, we observe a significant improvement on target class performance and generalized target performance compared to all ZSD baselines. This is mainly due to our ability to utilize additional, low-cost training data. We also outperform WSD and MSD baselines on target classes, despite their access to information about target classes during training, and we significantly outperform them on base classes and therefore overall, due to our effective exploitation of bounding box supervision for base classes. Note that WSD and MSD models cannot be evaluated on base-only or target-only classes since they have a fixed classifier trained on all 65 classes. Moreover, we have a FSD (fully supervised detection) baseline to measure the performance drop on base classes.

Furthermore, we present ablation experiments in Table~\ref{table:ablation_results} to show the effectiveness of each design choice. Particularly, we observe that without pretraining our model on image-caption datasets, the model performs poorly. This confirms the remarkable efficacy of multimodal pretraining for open-vocabulary generalization. We also observe that grounding is the main component of pretraining, which has a much larger effect than the auxiliary objectives that are optimized through the multimedia transformer module. Moreover, we show that transferring ResNet weights alone (from pretraining to downstream task) is not enough for effective knowledge transfer, and we must transfer the V2L layer as well. Additionally, if the V2L layer is not frozen during downstream training, it loses its ability to generalize to target classes, in order to slightly improve on base classes. 

We also try initializing the model randomly during pretraining instead of using widely used Imagenet weights, and despite the performance drop, we still perform better than most ZSD baselines that use Imagenet. We also observe that if we use the automatically collected Conceptual Captions instead of the carefully annotated COCO Captions, the performance drops, but still outperforms all ZSD baselines significantly, proving that even low-quality, cheap data can be utilized by OVR-CNN to achieve better performance. 

\begin{table}[tb]
\begin{center}
\begin{tabular}{l|c|c|c}
\hline
\multirow{2}{*}{Ablation} & Base & Target & All \\
& (48) & (17) & (65) \\ 
\hline
\hline
Ours w/o pretraining & 25.2 & 4.4 & 18.1 \\
\hline
Ours w/o grounding & 25.9 & 4.6 & 19.0 \\
Ours w/o auxiliary objectives & 45.6 & 26.0 & 38.8 \\
\hline
Ours w/o transferring V2L  & 25.3 & 4.9 & 18.6 \\
Ours w/o freezing V2L & 47.0 & 23.4 & 39.3\\
\hline
Ours w/o Imagenet & 18.4 & 9.13 & 14.3 \\
Ours w/ Conceptual Captions & 43.0 & 16.7 & 34.3 \\
\hline
Ours & \textbf{46.8} & \textbf{27.5} & \textbf{39.9}\\
\hline
\end{tabular}
\end{center}
\caption{Ablation on MSCOCO dataset. Numbers are mAP (\%).}
\label{table:ablation_results}
\vspace{-0.1cm}
\end{table}

\subsection{Visualization and discussion}

\begin{figure}[t]
\begin{center}
 \includegraphics[width=1.00\linewidth]{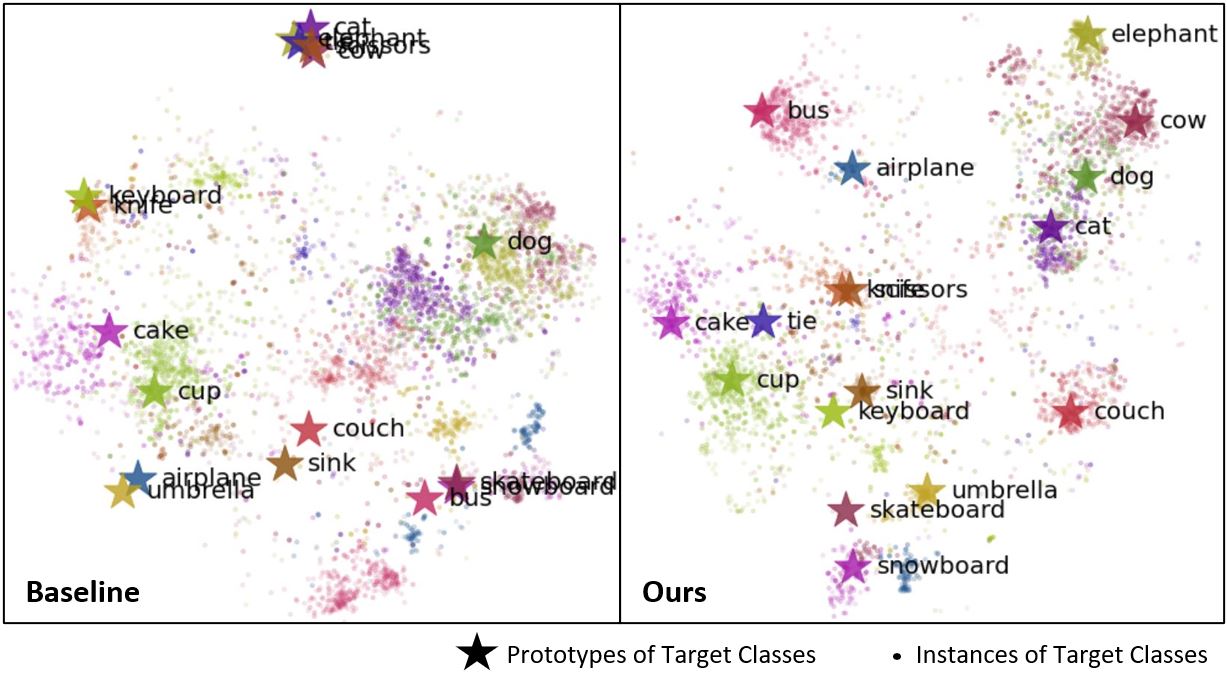}
\end{center}
\vspace{-0.1cm}
   \caption{The embedding space learned by OVR-CNN (right) compared to a baseline without pretraining (left). Each color represents a target class, each dot represents the $e^I_i$ embedding of a bounding box and each star represents a class prototype.}
\label{fig:tsne}
\vspace{-0.1cm}
\end{figure}

To gain deeper insight about what OVR-CNN learns, we depict the visual-semantic embedding space that is learned by our model in Figure~\ref{fig:tsne}. More specifically, we apply our trained model (after downstream training) on all COCO validation images, get the embeddings of all output bounding boxes after the V2L layer $e^I_i$, and reduce their dimensionality to 2 using t-SNE~\cite{vandermaaten08a}. We color-code them based on their ground truth label and overlay class embeddings $e_k^\mathcal{V}$ on the same space. We only show target classes and their instances to reduce clutter. Ideally, instances of each target class should form distinct clusters, and each class embedding (prototype) should fall inside the cluster formed by its instances. This is particularly difficult to achieve for target classes due to the lack of direct supervision. We compare our method to a ZSD baseline that is identical to our model except without pretraining on image-caption pairs. 

We observe that in the baseline, target classes form convoluted clusters and their prototypes are randomly distributed or collapsed. On the other hand, our full model creates well-defined clusters that contain their prototypes in most cases. This is consistent with our intuition and our quantitative results that suggest zero-shot learning is not sufficient for learning a smooth and generalizable mapping from visual features to semantic embeddings, and learning a larger vocabulary through multimodal data is crucial for a more coherent space and generalizing beyond base classes.

\section{Conclusion}
\label{sec:conclusion}

We called attention to the new task of Open-Vocabulary Object Detection (OVD), as an attempt to disentangle object detection into recognition and localization, and learn them separately using two different sources of supervision that are perfect for each corresponding task. In OVD, recognition is learned from captions, which are general-purpose and open-vocabulary, while localization is learned from bounding box annotations, which are accurate and directly designed for the downstream task. We proposed OVR-CNN which pretrains a Faster R-CNN on an image-caption dataset and carefully transfers the open-vocabulary visual-semantic knowledge learned from captions to the downstream task of object detection. We demonstrated record performance compared to zero-shot and weakly supervised baselines, establishing a new state of the art for scalable object detection. Nevertheless, OVR-CNN is merely one possible implementation of our general idea, which can be extended to other downstream tasks too, enabling more human-like, open-vocabulary computer vision technology. 


{\small
\bibliographystyle{ieee_fullname}
\bibliography{egbib}
}

\pagebreak
\section{Supplementary Material}

\begin{figure}[t]
\begin{center}
 \includegraphics[width=1.00\linewidth]{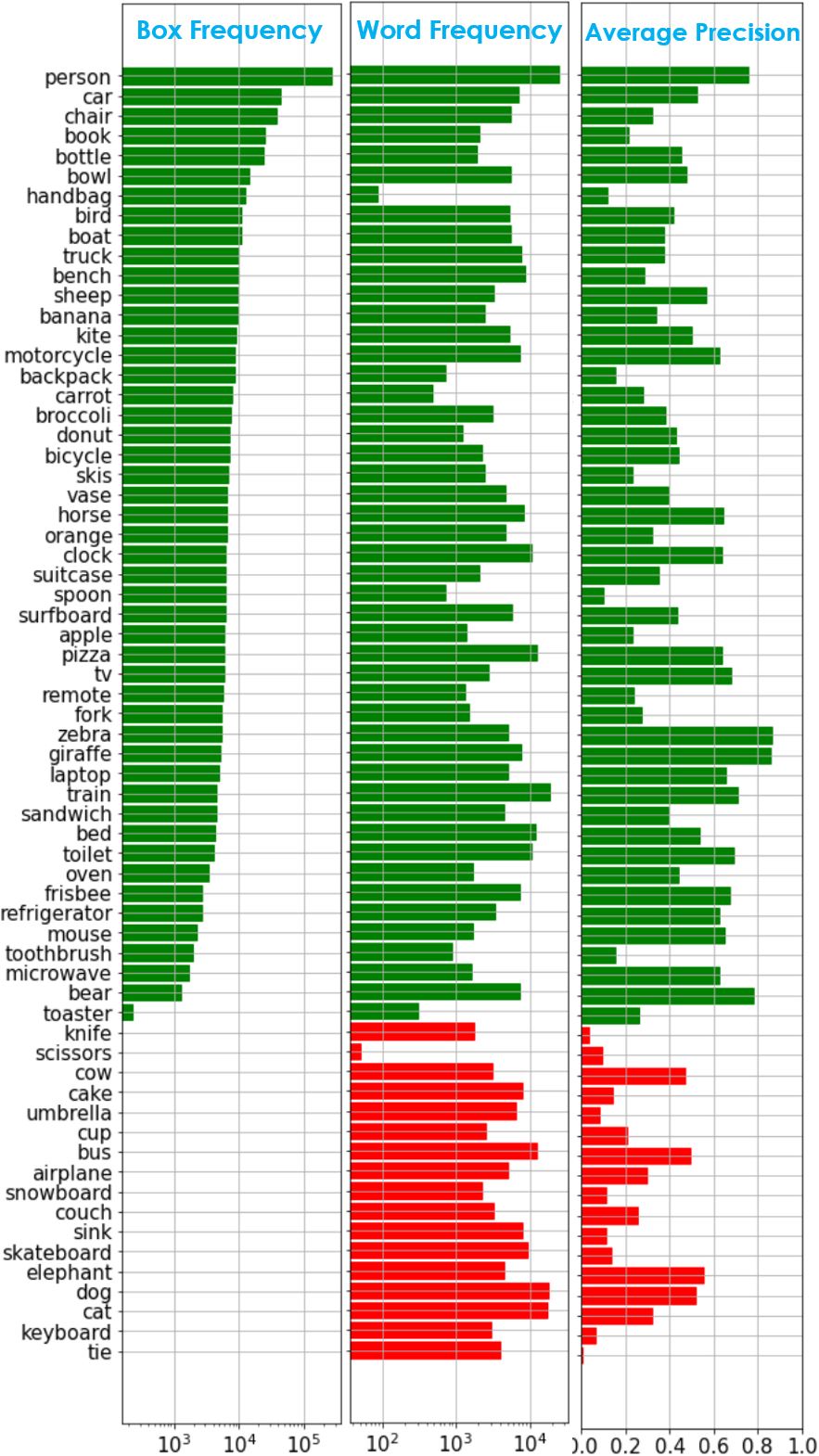}
\end{center}
   \caption{Performance for each class along with data frequency during pretraining and downstream training. Green and red show base and target classes respectively.}
\label{fig:perclass}
\end{figure}

In this section, we provide statistical and qualitative analysis to gain additional insights about the performance of the proposed method. Since one of the most critical issues of deep learning is bias, we start by analyzing the effect of training data bias on our per-class performance. Since we have two training phases, the class frequency during pretraining and downstream training should be separately analyzed. Figure~\ref{fig:perclass} shows our per-class performance (right), along with the frequency of bounding box instances during downstream training (left), and the frequency of words during pretraining (center). 

Our first observation is that our performance is not affected by the bias in downstream training data. As we move down the list, classes become exponentially less frequent, but the performance does not drop at all, except target (red) classes which have exactly zero examples during downstream training, and are inevitably less accurate. Our robustness to data bias is most likely due to the fact that we fix the classification head during downstream training, including both the V2L layer and the class embeddings. This is in contrast with conventional classifiers which fully adapt the classifier parameters, including an explicit \textit{bias} term, to the biased training data.

Nevertheless, when we compare the performance to word frequency during pretraining, we do observe a correlation between the least frequent words and the lest accurate classes. This correlation is not very strong, but it motivates our future work on bias mitigation mechanisms that can be used in naturally supervised (image-caption) settings. 

Furthermore, we observe that smaller objects such as \texttt{knife} and \texttt{tie} have lower performance, which is to some extent consistent with supervised object detection, but is fueled by the fact that our grounding mechanism is weakly supervised, and is less likely to correctly align smaller objects to words, because they take a smaller portion of the feature map. 

\begin{figure*}[t]
\begin{center}
 \includegraphics[width=1.00\linewidth]{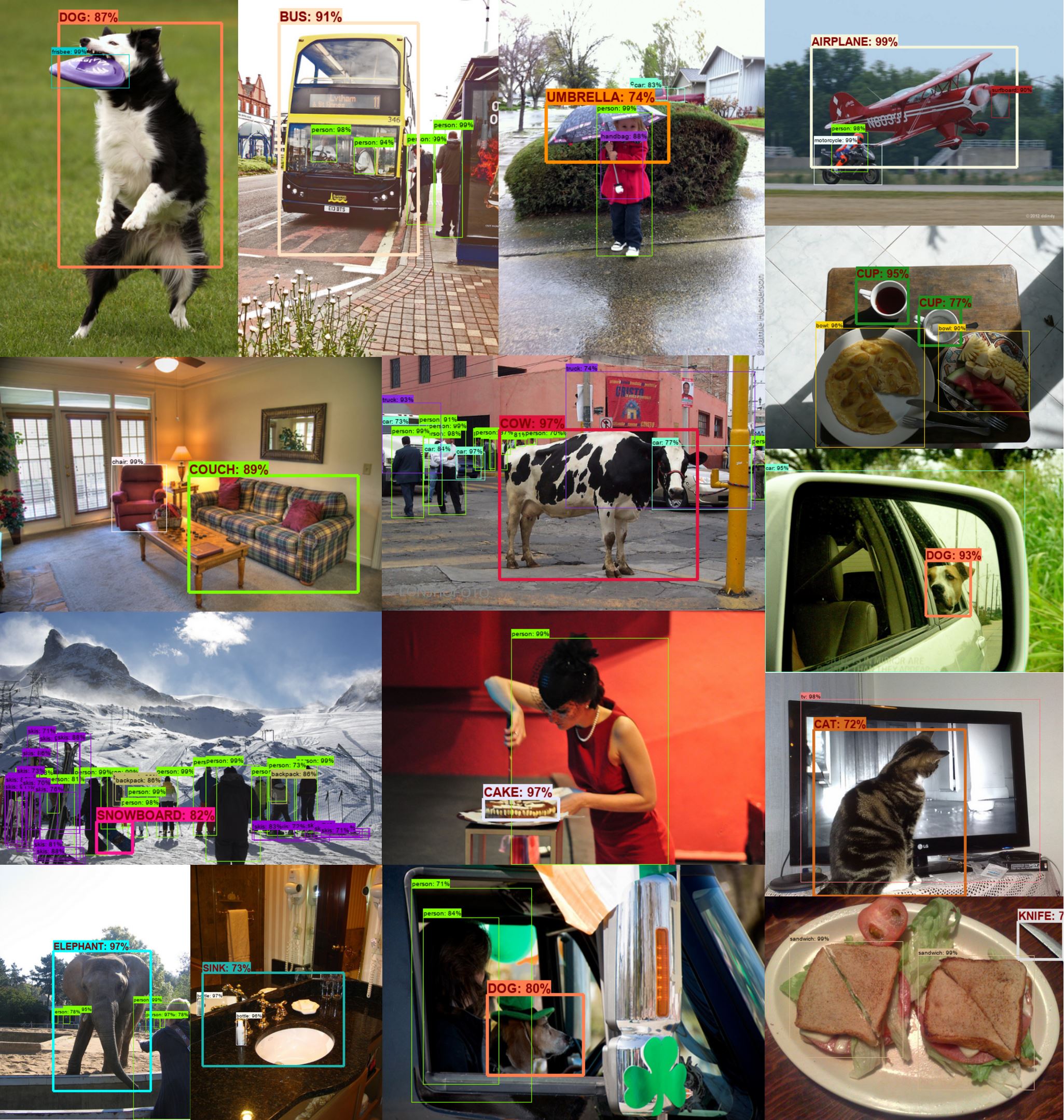}
\end{center}
   \caption{Qualitative results of our OVR-CNN model, detecting both base and target classes. Target classes are shown with larger font, thicker border, and uppercase.}
\label{fig:examples}
\end{figure*}

To get a qualitative look at the performance, we deploy our model on the COCO validation set and visualize its detection outputs in Figure~\ref{fig:examples}. We use the generalized version which selects the category of each object from the union of base and target classes. We emphasize target classes for better visibility, and analyse the quality of the predictions. Based on our observation, the main limitation of our method is localization accuracy for target classes. There are several cases of overly loose or overly tight bounding boxes, which is due to the fact that we have no ground truth bounding boxes for target classes. This motivates future work on class-agnostic boundary refinement.

\end{document}